\pdfoutput=1

\documentclass[11pt]{article}

\usepackage[]{emnlp2021}

\usepackage{times}
\usepackage{latexsym}
\usepackage{amssymb}
\usepackage{xspace}
\usepackage[T1]{fontenc}
\usepackage{graphics}
\usepackage[utf8]{inputenc}

\usepackage{microtype}
\usepackage{amsmath}
\title{Revisiting Automatic Question Summarization Evaluation in the Biomedical Domain}

\author{Hongyi Yuan$^{12}$\thanks{$\quad$Work done at Alibaba DAMO Academy.},\xspace\xspace Yaoyun Zhang$^1$\xspace\xspace, Fei Huang$^1$\xspace\xspace, Songfang Huang$^1$\\
        $^1$Alibaba Group, $^2$Tsinghua University\\
        \texttt{yuanhy20@mails.tsinghua.edu.cn}\\
        \texttt{\{zhangyaoyun.zyy,f.huang,songfang.hsf\}@alibaba-inc.com}}

\begin{document}
\maketitle
\begin{abstract}
Automatic evaluation metrics have been facilitating the rapid development of automatic summarization methods by providing instant and fair assessments of the quality of summaries. 
Most metrics have been developed for the general domain, especially news and meeting notes, or other language-generation tasks. 
However, these metrics are applied to evaluate summarization systems in different domains, such as biomedical question summarization.
To better understand whether commonly used evaluation metrics are capable of evaluating automatic summarization in the biomedical domain, we conduct human evaluations of summarization quality from four different aspects of a biomedical question summarization task.
Based on human judgments, we identify different noteworthy features for current automatic metrics and summarization systems as well.
We also release a dataset of our human annotations to aid the research of summarization evaluation metrics in the biomedical domain.
\end{abstract}

\section{Introduction}

Automatic summarization builds succinct summaries from long source texts. Since source texts may be corresponding to multiple good summaries, and considering only one reference summary for each text are given in most of the datasets, accurate assessment of summaries quality is a focused area of summarization research. \cite{gao-wan-2022-dialsummeval,kryscinski-etal-2019-neural}


A number of automatic evaluation metrics have been developed to provide an assessment of summarization quality and facilitate the rapid development of summarization algorithms. \cite{zhang20bertscore} These metrics are understandably developed and evaluated in the general domain task (i.e. news \cite{grusky-etal-2018-newsroom,cnndm}) for improving summarization algorithms. 

In general domain summarization tasks, especially news summarization, the performance of abstractive summarization is significantly improved with the recent success of pre-trained generative language models.~\cite{lewis-etal-2020-bart} However, the summarization in other domains is less studied. Domain-specific summarization tasks may present different challenges for summarization systems and automatic evaluation systems, such as the degree of conciseness and the need for in-domain knowledge. The metrics developed for general purposes may not be effective for specialized domains.

In this paper, we investigate the effectiveness of automatic summarization evaluation metrics in the biomedical domain. We revisit the existing metrics using a healthcare question summarization task. The main contributions are as follows:

$\bullet$ We use HealthCareMagic \cite{zeng-etal-2020-meddialog} as the biomedical question summarization datasets. We train different Transformers-based language models to produce the summarization. We recruit professional annotators who are fluent in English, and annotators for the task have biomedical experience backgrounds. We select a subset of test samples and evaluate summaries from different models by annotators.

$\bullet$ We use human annotations to assess the generated summary quality from different models and revisit the correlation of different metrics with human judgments. Analyses throw light on the features and challenges of this biomedical domain summarization task for summary generation and automatic evaluation. 

$\bullet$ We will release the annotation results for future research to aid the development and assessment of summarization metrics for biomedical tasks.



\section{Summary Collection}

\subsection{Data}

For biomedical question summarization, we use HealthCareMagic dataset, which contains samples from MedDialog \cite{zeng-etal-2020-meddialog} dataset. Each sample is comprised of a dialogue between a patient and a doctor inquiring about healthcare instruction and a short summary briefing the patient's question. Summaries are abstractive and are written in a formal style. Overall, HealthCareMagic contains 226,405 samples and we follow previous work \cite{mrini-etal-2021-gradually} for the size of training, validation, and testing set splits. This results in 181,122 training samples, 22,641 validation samples, and 22,641 testing samples. As the target texts are the summaries of patients' questions, we leave out the doctors' feedback from the source input tests when training and evaluation.



\subsection{Summary Generation Model}

Following~\cite{goodwin-etal-2020-flight}, we use the single systems of Pegasus~\cite{zhang2019pegasus}, BART~\cite{lewis-etal-2020-bart}, T5~\cite{raffel2020t5} and BioBART \cite{yuan-etal-2022-biobart} which continuously trains BART on PubMed abstracts with text infilling and sentence permutation to generative different question summaries. We leave the detailed introduction of systems to Appendix \ref{appC}.

We finetune the baseline models for 3 epochs HealthCareMagic dataset, with batch size 16, learning rate 5e-5, and an AdamW optimizer on Nvidia V100 GPUs. Our implementation is based on Huggingface \cite{wolf-etal-2020-transformers}, and we use the large version of all four models in our experiments.


\section{Human Annotation}

Since the test set size of HealthCareMagic is large and human annotating is time-consuming and expensive, we randomly select a subset of 500 samples for human annotation to avoid a large annotation workload. The average source question length is 77.94 words. Figure \ref{fig:sample_length} in Appendix \ref{appA} depicts the input word length distribution of selected samples.

Following previous research \cite{gao-wan-2022-dialsummeval,kryscinski-etal-2019-neural}, we demand human annotators evaluate samples on the \textit{summary level} from the following three aspect:

\textbf{Relevance} measures how well the question summary captures the main concerns of the patient’s questions. This focuses on the coverage of the summary for the corresponding questions.

\textbf{Consistency} measures whether the question summary is consistent with the patients’ question. A good summary may accurately capture the main concerns of summaries and may not contain untrue and hallucinatory information.

\textbf{Fluency} measures the grammatical and syntactic quality of the summary sentences.

Besides, we also consider an additional overall evaluation aspect on the \textit{sample level}. 

\textbf{Speciality} gives an overall rating for each sample indicating whether the special knowledge (biomedical domain) is required for generating good summarization of patients' questions. The annotators will assess the content of the patents’ question and all the summaries en bloc.

For each of the aspects, we demand the annotators to give an integer rating range from 1 to 5 for each sample, where 5 for the best. The workload is to give $500$(Samples Num.)$\times 5$(Summary Num. per Sample)$\times 3$(Aspect Num.)=$7500$ summary ratings and $500$(Samples Num.) sample ratings. 

During the annotation process, we assign two human annotators for separated evaluation. To control the annotation quality, after the separated annotation, a third-party annotator is demanded to revise both human ratings. The third-party annotator then gives final ratings for each sample or summary.




\begin{table}
\centering
\begin{tabular}{lcc}
&\textit{Separated} & \textit{Revised} \\
\hline
\textbf{Relevence} &56.85&72.81 \\
\textbf{Consistency} &74.82& 86.65 \\
\textbf{Fluency} &67.54&82.19 \\
\textbf{Speciality} &64.77&80.78 \\
\hline
\end{tabular}
\caption{\label{human-agreement}
Inter-annotator agreements by Cohen's $\kappa$ for separated and revised human annotations.
}
\end{table}

\section{Automatic Evaluation Metric}

We give brief introductions of selected automatic language generation metrics. The following metrics compare the n-gram features of generated summaries and reference summaries.

\textbf{BLEU}~\cite{papineni-etal-2002-bleu} is a precision-oriented metric based on n-gram matching. BLEU is a popular metric for machine translation tasks. Short hypotheses are penalized with a brevity penalty. We also consider a BLEU variant, SacreBLEU \cite{post-2018-call}, since BLEU scores are usually biased by different word tokenizers.

\textbf{Rouge}~\cite{lin-2004-rouge} is a well-known metric for summarization which is based on n-gram matches (Rouge1/2) or longest common sub-sequence (RougeL). The F1 score in Rouge is commonly weighted to favor recall over precision. 

\textbf{METEOR}~\cite{lavie-agarwal-2007-meteor} introduces stem, word orders, and synonyms from WordNet beyond n-grams for evaluation. 

\textbf{CHRF}~\cite{popovic-2015-chrf} evaluates generated summaries based on char-level n-gram F1 score between generated and reference summaries. 

Since n-gram-based metrics may neglect in-depth semantic features, automatic metrics are proposed based on pre-trained language models (PLMs). 

\textbf{BertScore}~\cite{zhang20bertscore} utilizes PLMs to generate contextual embeddings for source and target texts and scores the generated and reference pairs based on the alignment between embeddings. Since the dataset is derived from the biomedical domain, we use the general domain and scientific domain PLMs, RoBERTa \cite{Liu2019RoBERTaAR} and SciBERT \cite{beltagy-etal-2019-scibert}, to generate embeddings.

\textbf{BLANC}~\cite{vasilyev-etal-2020-fill} is a reference-free metric. It compares BERT's \cite{devlin-etal-2019-bert} performance difference on masked token reconstruction with and without summaries to evaluate. We use BLANC-help and BLANC-tune. 

\textbf{Perplexity} (PPL), is widely used for evaluating the quality of language fluency in the language modeling task \cite{attention}. PPL is based on the token-level likelihood generated by PLMs and is free of source questions and reference summaries during evaluation. We use GPT-2 \cite{radford2019language} to compute PPL for each summary. 

\textbf{MAUVE}~\cite{pillutla2021mauve} evaluates the generated summaries from a distributional perspective. The distribution discrepancy between real and generated texts is measured by KL-divergence of GPT-2 derived likelihoods. 




\section{Results}

\subsection{Human Annotation Agreement}

We evaluate the inter-annotation agreements via Cohen’s $\kappa$ \cite{McHugh2012InterraterRT}. The evaluation results are listed in Table \ref{human-agreement}. For the separated annotations, the inter-annotator agreements are 56.85, 74.82, and 67.54 in terms of Relevance, Consistency, and Fluency respectively on summary-level evaluation. On sample level evaluation, the agreement on Speciality is 64.77. For revised annotations, the agreements on Relevance, Consistency, and Fluency are increased to 72.81, 86.65, and 82.19, and on Speciality the agreement score rises to 80.78. The agreement scores are within acceptable ranges. Even after third-party revision, there is still a discrepancy in perfection. This shows that language evaluation is hard and controversial. 

\subsection{Model Assessment}

\begin{table}[t]
    \centering
    \resizebox{.45\textwidth}{!}{
    \begin{tabular}{lcc}
 & Annotator1 & Annotator2 \\
\hline
&\multicolumn{2}{c}{\textbf{Relevance/Consistency/Fluency}} \\
Reference  & 3.61/3.27/4.74 & 3.67/3.26/4.74 \\
BART  & 3.61/3.22/4.69 & 3.73/3.24/4.72 \\
BioBART  & 3.58/3.13/4.71 & 3.69/3.15/4.74 \\
T5 & 3.57/3.16/4.73 & 3.69/3.19/4.77 \\
PEGASUS & 3.61/3.21/4.81 & 3.72/3.24/4.83 \\
\hline
\hline
\textbf{Speciality}& 1.53 & 1.56 \\
\hline
\end{tabular}
}
\caption{The averaged summary level annotation scores for each summarization model in terms of Relevance, Consistency, and Fluency. The last line is the sample level Speciality score from different annotators.}
\label{tab:model_assess}
\end{table}

\begin{table*}[t]
    \centering
    \resizebox{2.0\columnwidth}{!}{
    \begin{tabular}{l|cc|cc|cc}
    \multicolumn{1}{c}{}&\multicolumn{2}{c}{Relevance} &\multicolumn{2}{c}{Consistency}&\multicolumn{2}{c}{Fluency}\\
    \hline
    Metric&Annotator 1&Annotator 2&Annotator 1&Annotator 2&Annotator 1&Annotator 2\\
    \hline
    \multicolumn{7}{c}{\textit{n-gram based metric}}\\
    SacreBLEU&16.18&13.45&17.59&19.11&6.08&4.98\\
    BLEU&10.59&7.44&12.45&13.58&3.28&1.60\\
    Rouge1&18.79&19.36&17.58&19.93&-2.87&-1.88\\
    Rouge2&16.46&15.57&15.46&17.22&-1.93&-1.93\\
    RougeL&18.00&18.52&17.90&20.10&-1.66&-0.31\\
    METEOR&20.76&20.18&22.79&24.85&1.47&1.95\\
    Crhf&20.99&20.86&24.00&25.32&-1.33&-1.08\\
    \hline
    \multicolumn{7}{c}{\textit{PLM based metric}}\\
    BS-SciBERT&17.70&16.99&19.32&21.28&7.18&8.50\\
    BS-RoBERTa&19.89&18.77&20.89&22.24&10.59&13.24\\
    BLANC-help&7.68&9.16&10.71&9.28&-7.72&-7.36\\
    BLANC-tune&6.63&8.54&6.57&6.84&-9.13&-9.89\\
    PPL&-8.37&-4.99&-6.53&-3.20&-0.52&-0.74\\
    MAUVE&1.55&0.43&0.67&0.42&1.56&1.24\\
    \end{tabular}
    }
    \caption{Correlations for metrics with human annotators. BS stands for BERTScore. PPL stands for perplexity.}
    \label{tab:metric-assess}
\end{table*}

Since the revised annotations are of higher agreement, we use the revised ratings to illustrate the models’ performance. \footnote{We leave the separated annotation results in Appendix \ref{appB}.} As shown in Table \ref{tab:model_assess}, in terms of summary level evaluation, the reference summaries do not show superiority over the generated summaries. The summaries from different language models are of similar quality. For different aspects, summaries from all sources achieve high Fluency, this indicates that it is relatively easy to generate grammatically correct summaries which are short in this task. Generating relevant and consistent summaries with the source questions is the main challenge of this task.  

In terms of sample level evaluation, we can surprisingly find that the Speciality ratings are rather low. This shows that generating a good question summary for medical question narratives does not necessarily need special in-domain knowledge. The finding is consistent with that BioBART does not achieve superior performance over BART, while BioBART surpasses BART in many biomedical domain tasks. \cite{yuan-etal-2022-biobart}

\subsection{Metric Assessment}

We investigate the correlation between different automatic metrics and human ratings. The correlation scores are calculated as follows, for an automatic metric $M$ and a aspect of summary level ratings $A$, the correlation $\rho_{MA}$ is:
\begin{align*}
    \rho_{MA} = r\left([\dots,M(s_{ij}),\dots], [\dots,A(s_{ij}),\dots]\}\right),
\end{align*}
where $i$ and $j$ represent the indices of $N$ samples and $P$ summaries for each sample, and $r(\cdot,\cdot)$ is the correlation function.

The results are listed in Table \ref{tab:metric-assess}. (1) In the overview, the correlations between metrics and human annotations are pretty low. The best correlation by CHRF on Consistency is only 25.32, while PPL even achieves a negative correlation across all three aspects. (2) Comparing metrics' performance on three aspects, Fluency is received the low correlation scores, and this could be due to all the summaries achieving similarly good fluency which is consistent with the findings above. (3) Surprisingly, PLM-based metrics are not consistently superior to n-gram-based metrics. BLANC, PPL and MAUVE have low correlations with human judgments. This shows that n-gram-based metric is better suited for medical question summarization tasks in which summaries are short, and PLM-based metrics do not fit the task well. (4) The choice of PLMs influence the performance of metric. Using RoBERTa has better results than using SciBERT in BERTScore.

\section{Related work}
The rapid development of neural text generation models has brought increasing attention to assessing text evaluation metrics. \citet{bhandari_re-evaluating_2020} assessed the reliability of automatic metrics using both abstractive and extractive summaries. 
A dataset of collected human judgments was released.
Similarly, \citet{fabbri_summeval_2021} re-evaluated multiple automatic metrics for summarization and released a unified evaluation toolkit and an assembled collection of summaries. 
\citet{deutsch_re-examining_2022} identified the need to collect high-quality human judgments and improve automatic metrics when models' performance difference was marginal. 
However, these works mainly used news samples from CNN/Daily Mail for evaluation ~\cite{chen_thorough_2016}.

\citet{roy_reassessing_2021} extended such analyses to code summarization, and found 
that metric improvement scores of less than 2 points did not guarantee systematic improvements in summary quality.
\citet{gao-wan-2022-dialsummeval} revisited the evaluation of dialogue summarization and identified the correlation of automatic metrics with four aspects of human judgments. A dataset of human annotations for summaries was released.

\section{Conclusions and Future Work}

In summary, we report a pilot study evaluating different summarization models and assessing the effectiveness of automated summary evaluation metrics based on a biomedical domain question summarization task. The main findings show that relevance and consistency are the key obstacles. Summaries are of good fluency quality. We surprisingly find although the datasets are healthcare questions, they do need not much biomedical domain knowledge. PLM-based metrics do not uniformly surpass n-gram-based metrics, and all the metrics do not achieve a high correlation with human annotators. Our findings may enlighten potential direction for biomedical summarization research.



\section*{Limitations}

In this work, we investigate the current evaluation metrics for summarization in the biomedical domain. We select the most representative summarization models and evaluation metrics for evaluation, and we do not cover all the models and metrics due to paper length and human annotation workload. Although we intend to investigate metrics' performance in the biomedical domain. However, the annotation results in Speciality show that the selected dataset needs not much in-domain knowledge to generate a good question summary. This may be a potential limitation of the study.

\section*{Ethical Statements}

The human annotators are recruited university students with biomedical backgrounds. Human annotators take part in the annotation procedure through an online platform anonymously. Annotators are paid with four dollar per sample.

\bibliography{anthology,custom}
\bibliographystyle{acl_natbib}
\newpage
\appendix

\section{Data Distribution}
\label{appA}
\begin{figure}
    \centering
    \resizebox{.4\textwidth}{!}{
    \includegraphics{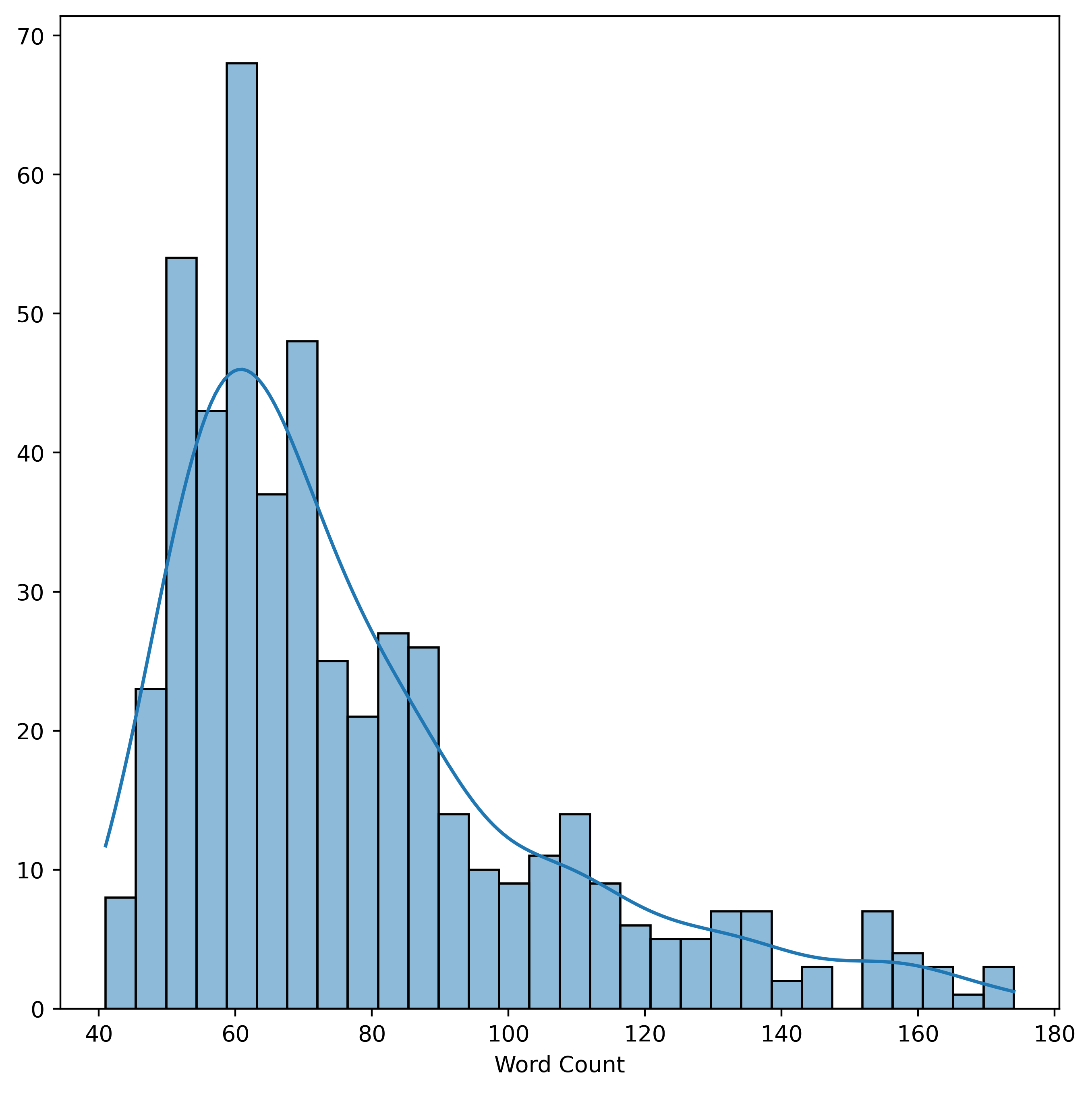}
    }
    \small
    \caption{The source question word length histogram of selected samples for human annotation.}
    \label{fig:sample_length}
\end{figure}

\section{Separated Annotation Results}
\label{appB}

\begin{table}[t]
    \centering
    \begin{tabular}{lcc}
 & Annotator1 & Annotator2 \\
\hline
&\multicolumn{2}{c}{\textbf{Relevance/Consistency/Fluency}} \\
Reference  & 3.55/3.24/4.73 & 3.69/3.24/4.74 \\
BART  & 3.55/3.18/4.67 & 3.73/3.24/4.72 \\
BioBART  & 3.51/3.11/4.68 & 3.70/3.14/4.75 \\
T5 & 3.49/3.13/4.70 & 3.70/3.19/4.77 \\
PEGASUS & 3.54/3.16/4.79 & 3.73/3.23/4.83 \\
\hline
\hline
\textbf{Speciality}& 1.49 & 1.55 \\
\hline
\end{tabular}
\caption{The averaged summary level annotation scores of separated annotation for each summarization model in terms of Relevance, Consistency, and Fluency. The last line is the sample level Speciality score showing whether generating good summaries for the sample requires in-domain knowledge. }
\label{app:model_assess}
\end{table}

\section{Introductions of Summarization Models }
\label{appC}

\textbf{Pegasus}~\cite{zhang2019pegasus} is a conditional language model designed specifically for abstractive summarization and pre-trained with a self-supervised gap-sentence-generation objective. Namely, the model is pre-trained to predict entirely masked sentences from the document.\\
\textbf{BART}~\cite{lewis-etal-2020-bart} is a generative pretrained language model that uses an encoder-decoder Transformers architecture and is pretrained to reconstruct text from several corruption noises. \\
\textbf{BioBART} is a generative language model adapted to the biomedical domain, which continuously trains BART on 41GB PubMed abstracts corpora with text infilling and sentence permutation since our work focuses on the summary generation of the biomedical domain. \\
\textbf{T5}~\cite{raffel2020t5} is pre-trained on multiple objectives, including masking, translation, classification, machine reading comprehension (MRC), and summarization, all formulated as conditional generation tasks.

\section{Data License}

The data is publicly availible on GitHub repository \textit{https://github.com/UCSD-AI4H/Medical-Dialogue-System.git}. The collector of the data did not clarify the license of the data. All copyrights of the data belong to healthcaremagic.com and icliniq.com.

\section{Pre-trained Model Checkpoints}

We use the model checkpoints on Huggingface:
\begin{enumerate}
    \item T5: {\small\texttt{t5-large}}
    \item BART: {\small\texttt{facebook/bart-large}}
    \item PEGASUS: {\small\texttt{google/pegasus-large}}
    \item BioBART: {\small\texttt{GanjinZero/biobart-large}}
\end{enumerate}
For each model, the training time is approximately 5 hours on one NVIDIA V100 GPU.

\begin{table*}[t]
    \centering
    \resizebox{2.0\columnwidth}{!}{
    \begin{tabular}{l|cc|cc|cc}
    \multicolumn{1}{c}{}&\multicolumn{2}{c}{Relevance} &\multicolumn{2}{c}{Consistency}&\multicolumn{2}{c}{Fluency}\\
    \hline
    Metric&Annotator 1&Annotator 2&Annotator 1&Annotator 2&Annotator 1&Annotator 2\\
    \hline
    \multicolumn{7}{c}{\textit{n-gram based metric}}\\
    SacreBLEU&13.62&14.20&15.88&18.84&4.44&4.59\\
    BLEU&9.06&8.23&10.99&13.66&2.04&1.22\\
    Rouge1&16.88&20.54&16.49&19.95&-4.74&-2.21\\
    Rouge2&14.86&16.71&14.46&17.11&-2.11&-2.15\\
    RougeL&16.31&19.46&17.01&19.93&-3.05&-0.58\\
    METEOR&19.37&20.98&21.77&24.52&-0.87&1.56\\
    CRHF&20.48&21.62&22.64&24.98&-2.43&-1.38\\
    \hline
    \multicolumn{7}{c}{\textit{PLM based metric}}\\
    BS-SciBERT&16.43&17.55&18.26&20.84&5.49&8.11\\
    BS-RoBERTa&17.82&19.16&19.13&22.05&8.93&12.94\\
    BLANC-help&6.50&8.57&10.32&8.81&-7.59&-7.29\\
    BLANC-tune&5.80&8.50&7.19&7.21&-10.29&-10.41\\
    PPL&-7.33&-4.28&-6.09&-2.04&-0.63&-0.39\\
    MAUVE&0.96&0.36&0.99&0.49&0.44&1.23\\
    \end{tabular}
    }
    \caption{Correlations for metrics with separated annotations. BS stands for BERTScore. PPL stands for perplexity.}
    \label{tab:metric-assess-separated}
\end{table*}



\end{document}